%% file: main.tex
\documentclass[10pt,twocolumn,letterpaper]{article}
\usepackage{threeparttable}
\usepackage{adjustbox}
\usepackage{multirow}
\usepackage{makecell}
\usepackage{subcaption}
\usepackage{soul}
\usepackage{pifont}
\usepackage{bbding}

\usepackage{cvpr}              
\newcommand{\bs}[1]{\boldsymbol{#1}}
\newcommand{\mr}[1]{\mathrm{#1}}

\definecolor{cvprblue}{rgb}{0.21,0.49,0.74}
\usepackage[pagebackref,breaklinks,colorlinks,allcolors=cvprblue]{hyperref}

\def\paperID{11733} 
\def\confName{CVPR}
\def\confYear{2025}

\title{Alignment, Mining and Fusion: Representation Alignment with Hard Negative Mining and Selective Knowledge Fusion for Medical Visual Question Answering}

\author{Yuanhao Zou\\
University of Michigan, Ann Arbor\\
{\tt\small yuanhaoz@umich.edu}
\and
Zhaozheng Yin\thanks{Corresponding author}\\
Stony Brook University\\
{\tt\small zyin@cs.stonybrook.edu}
}

\begin{document}
\maketitle
\input{sec/0_abstract}    
\input{sec/1_Intro}

\input{sec/2_Related}
\input{sec/3_Method}
\input{sec/4_Experiment}
\input{sec/5_Conclusion}
{
    \small
    \bibliographystyle{ieeenat_fullname}
    \bibliography{main}
}

\input{sec/X_suppl}

\end{document}

%% file: sec/0_abstract.tex
\begin{abstract}
Medical Visual Question Answering (Med-VQA) is a challenging task that requires a deep understanding of both medical images and textual questions. Although recent works leveraging Medical Vision-Language Pre-training (Med-VLP) have shown strong performance on the Med-VQA task, there is still no unified solution for modality alignment, and the issue of hard negatives remains under-explored. Additionally, commonly used knowledge fusion techniques for Med-VQA may introduce irrelevant information. In this work, we propose a framework to address these challenges through three key contributions: (1) a unified solution for heterogeneous modality alignments across multiple levels, modalities, views, and stages, leveraging methods like contrastive learning and optimal transport theory; (2) a hard negative mining method that employs soft labels for multi-modality alignments and enforces the hard negative pair discrimination; and (3) a Gated Cross-Attention Module for Med-VQA that integrates the answer vocabulary as prior knowledge and selects relevant information from it. Our framework outperforms the previous state-of-the-art on widely used Med-VQA datasets like RAD-VQA, SLAKE, PathVQA and VQA-2019. The code is available at \url{https://github.com/AlexCo1d/AMiF}
\end{abstract}

%% file: sec/1_Intro.tex
\section{Introduction}
\label{sec:intro}
\begin{figure*}[t]
  \centering
   \includegraphics[width=1\textwidth]{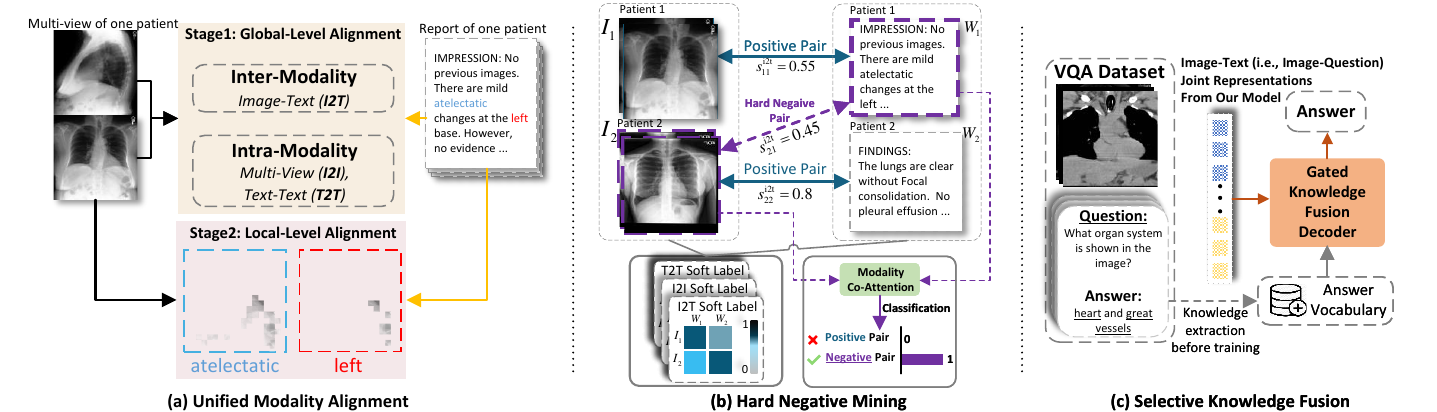}

   \caption{Motivations on: (a) unified modality alignment, (b) hard negative mining, and (c) selective knowledge fusion for Med-VQA.}
   \label{fig:overview}
\end{figure*}

Medical Visual Question Answering (\textbf{Med-VQA}) is a task that requires a joint understanding of both medical texts and images, aiding in clinical decision-making \cite{lin2023medvqa-survey}. While deep learning models have made significant strides in general visual question answering tasks \cite{manmadhan2020vqa-survey}, Med-VQA presents unique challenges, such as the presence of obscure objects in images and relatively smaller scale of annotated medical datasets for training models \cite{lapa, tanida2023interactive}. To address these challenges, many existing works \cite{gloria, mgca, mumc, m2i2, m3ae, arl, pubmedclip} utilize Vision-Language Pre-training (\textbf{VLP}), achieving remarkable success in the medical domain (\ie, Medical Vision-Language Pre-training, \textbf{Med-VLP}). These methods employ self-supervised learning to pre-train models on large unannotated medical datasets (\eg, MIMIC-CXR \cite{data-mimic}), initializing their models with the ability to extract medical visual-language representations. The models are then fine-tuned for downstream tasks like Med-VQA.

Among Med-VLP research, self-supervised techniques such as Contrastive Learning \cite{clip, simclr, convirt} are widely used to train visual and textual encoders to align representations between images and texts. While some works \cite{convirt, pubmedclip, zhang2023biomedclip, mumc, cmitm, mrm-iclr, medclip, m2i2} perform the global-level alignment (\ie, align only the \texttt{[CLS]} token of images and texts), other works \cite{gloria,lovt,mgca,dawidowicz2023limitr} attempt the local-level alignment, matching the similarity between image tokens and text tokens. However, these works focus on the inter-modality information (aligning images with texts), ignoring that medical datasets for Med-VLP (\eg, MIMIC-CXR \cite{data-mimic}) contain valuable intra-modality similarities within images and reports. Though works like MUMC \cite{mumc} attempt to align features within uni-modality, they do not perform local-level alignment and overlook the multi-view attribute of radiology imaging, which can naturally be used for intra-modality alignment. \ul{A unified approach that simultaneously addresses global-level alignment for both inter-modality (image-text) and intra-modality (image-image \& text-text), and local-level alignment remains under-explored}.

Additionally, contrastive learning methods like CLIP \cite{clip} treat image-text \textit{pairs} (image-text from the same patient) as positive samples, with unpaired images and texts as negative samples. However, in the medical domain, because of similar clinical manifestations of the same diseases, consistent anatomical imaging patterns, and standardized reporting practices, high similarities exist between unpaired images and texts, resulting in numerous hard negatives. While most of the prior CLIP-based works \cite{convirt, pubmedclip, zhang2023biomedclip, gloria, mgca, m2i2, cmitm, mrm-iclr} employ one-hot labels for contrastive learning, they overlook the issue of hard negatives in the medical domain. Though prior works \cite{medclip, reco, mumc, chen2023kobo-rethink-medvlp} use soft labels for hard negatives, their methods primarily consider hard negatives derived from image-text similarity (\ie, inter-modality), neglecting that the tight relationships within images and reports (\ie, intra-modality) could also be leveraged to explore hard negatives. Moreover, these soft label approaches calculate non-zero or even high similarity scores for hard negative pairs, making it hard to explicitly distinguish them from positive pairs. \ul{Thus, \textit{hard negative mining}, \ie, distinguishing hard negative pairs from positive pairs for both inter-modality and intra-modality could be significantly improved in the medical domain.}

For fine-tuning downstream tasks like Med-VQA, fusing prior knowledge with the input image-question representation is a common practice to enhance the model's question-answering capability \cite{arl,lapa, data-slake, peters2019knowledge, wang2021kepler,nguyen2019mevf,mmq2021}. For example, many methods introduce knowledge graph matrices from Unified Medical Language System (UMLS, \cite{bodenreider2004umls}) as prior knowledge. This prior knowledge contains terminology that covers a comprehensive scope of the medical domain, while a downstream task may only focus on a specific medical imaging modality or a specific human organ. Therefore, fusing untrimmed knowledge may introduce irrelevant or unnecessary information. \ul{For Med-VQA, a method that can selectively fuse appropriate prior knowledge with the joint image-question representation will lead to a better question-answering ability for the model}. 

In this paper, to unify heterogeneous modality alignments and address the challenges of hard negatives and knowledge fusion in Medical Visual Question Answering, we propose a systematic framework named \textbf{AMiF}: Representation \textbf{A}lignment with Hard Negative \textbf{Mi}ning and Selective Knowledge \textbf{F}usion. Our detailed contributions are:

\begin{itemize}
    \item We unify various modality alignments (\cref{fig:overview} (a)) for medical representation learning at \textbf{Multi-level}: we use contrastive learning method for global-level alignment and optimal transport theory for local-level alignment; \textbf{Multi-modality}:  we align not only the inter-modality (I2T: image-text) but also the intra-modality (I2I: image-image, T2T: text-text); \textbf{Multi-view}: we leverage the unique multi-view attribute in the medical datasets for the intra-modality I2I alignment; and \textbf{Multi-stage}: we focus on the global-level alignment in the first stage of pre-training, and then focus on local-level alignment and hard negative mining in the second stage.

    \item We propose a hard negative mining method (\cref{fig:overview} (b)) that operates across multiple modalities during pre-training. Specifically, for both inter-modality and intra-modality alignments, we employ soft labels that assign non-zero values to negative pairs, which reflect the nuanced similarities between their representations due to similar disease characteristics. Subsequently, hard negative pairs mistakenly with high similarities are enforced to be identified and differentiated from positive pairs, helping AMiF learn discriminative image-text joint feature representations.

    \item For the Med-VQA task (\cref{fig:overview} (c)), we extract the answer vocabulary from the task-specific dataset as prior knowledge. We introduce a Gated Cross-Attention Module to fuse the prior knowledge into the image-text joint representation. Then, the gated operation in the module selects useful information from the knowledge-fused representation to enhance model performance on VQA tasks.
    
    \item By applying the above techniques, our proposed AMiF shows its effectiveness on the Med-VQA tasks, evaluated on widely-used benchmark datasets including RAD-VQA \cite{data-vqarad}, SLAKE \cite{data-slake}, PathVQA \cite{data-pathvqa} and VQA-2019 \cite{data-vqa2019}.
    
\end{itemize}

%% file: sec/2_Related.tex
\section{Related Works}
\label{sec:related works}

\subsection{Modality Alignment for Med-VLP}
While global-level alignment \cite{convirt, pubmedclip, zhang2023biomedclip, mumc, cmitm, mrm-iclr, medclip, m2i2} aligns the high-level \texttt{[CLS]} token of the modality representations, local-level alignment attempts to match all tokens of the representations. Numerous works have investigated the local-level alignment in the medical domain. Notably, Liao \etal compute the mutual information between local tokens \cite{liao2021local-mutual-information}, while GLoRIA \cite{gloria}, LRCLR \cite{rizvi2023lrclr}, LoVT \cite{lovt} and LIMITR \cite{dawidowicz2023limitr} explore using contrastive learning to align image patch tokens and report tokens at local-level \cite{Shrestha2023-medvlp-review}. MGCA \cite{mgca} uses a cross-attention module for local alignment. Unlike the idea of contrastive learning, UNITER \cite{uniter} and ViLT \cite{vilt} apply the Optimal Transport (OT) theory and IPOT algorithm \cite{ipot-algorithm} to optimize a transport plan for local-level alignment in the natural domain. The OT theory has not been used for Med-VLP yet. For multi-modality alignment, MUMC \cite{mumc} introduces uni-modality alignment within images or texts. While LIMITR \cite{dawidowicz2023limitr} compares both the frontal view and lateral view with the text representations. Our approach, which also explores multi-view images, focuses on alignment within the multi-view representations and unifies heterogeneous local-level alignment and global-level alignment on both inter- and intra-modalities.

\subsection{Hard Negatives for Med-VLP} 
For Med-VLP, CLIP-based methods use the one-hot label for image-text pairs \cite{convirt, pubmedclip, zhang2023biomedclip, mumc, cmitm, mrm-iclr, medclip, m2i2}, ignoring the hard negatives caused by the similarity between unpaired image-text representations in the medical domain. Many efforts attempt to solve the hard negative problem by applying soft label supervision for contrastive learning \cite{huang2024cusa, andonian2022progressive-self-distillation,albef, unilm, radenovic2023Filtering-distillation-hard-negatives,liu2021cprd}. In the medical domain, Jang \etal relax the similarity function by clipping the upper bound \cite{jang2024significantly}, ReCO \cite{reco} applies less penalty for the negative pairs, while MedCLIP \cite{medclip} and KoBo \cite{chen2023kobo-rethink-medvlp} use soft label from external knowledge. In addition, MUMC \cite{mumc} uses the self-distillation technique \cite{albef} to learn from the soft label generated itself. Different from the above studies, our approach explores hard negative mining in the medical domain. We not only use soft labels for hard-negative pairs of both inter-modality and intra-modality alignments but also attempt to further distinguish the hard negatives from the positives.

\subsection{Med-VQA and Knowledge Fusion}
For Med-VQA, prior works like MEVF \cite{nguyen2019mevf} and MMQ \cite{mmq2021} attempt to train a meta-model using additional data besides the Med-VQA dataset to facilitate VQA performance. Other methods \cite{m3ae,arl,mumc,m2i2,pubmedclip,zhang2023biomedclip} employ Med-VLP to initialize the model weight on large unlabeled datasets via self-supervised learning. Besides, without pre-training, other approaches introduce structured prior knowledge. For instance, LaPA \cite{lapa} utilizes a knowledge graph \cite{velivckovic2017graph-attention-network} with well-stored medical information as its prior knowledge, integrating auxiliary information for more relevant answer prediction. Our approach uses answer vocabulary from the specific Med-VQA datasets as prior knowledge and enables selective knowledge fusion via a Gated Cross-Attention Module. Moreover, most prior works \cite{m3ae,lapa,m2i2,arl} treat Med-VQA as a multi-labels classification task, using a fixed number of answers for prediction, while our model generates \textit{open}-form answers like \cite{albef,q2atransformer}.

%% file: sec/3_Method.tex
\section{Methodology}
\label{method}

\begin{figure*}[t]
  \centering
   \includegraphics[width=1\textwidth]{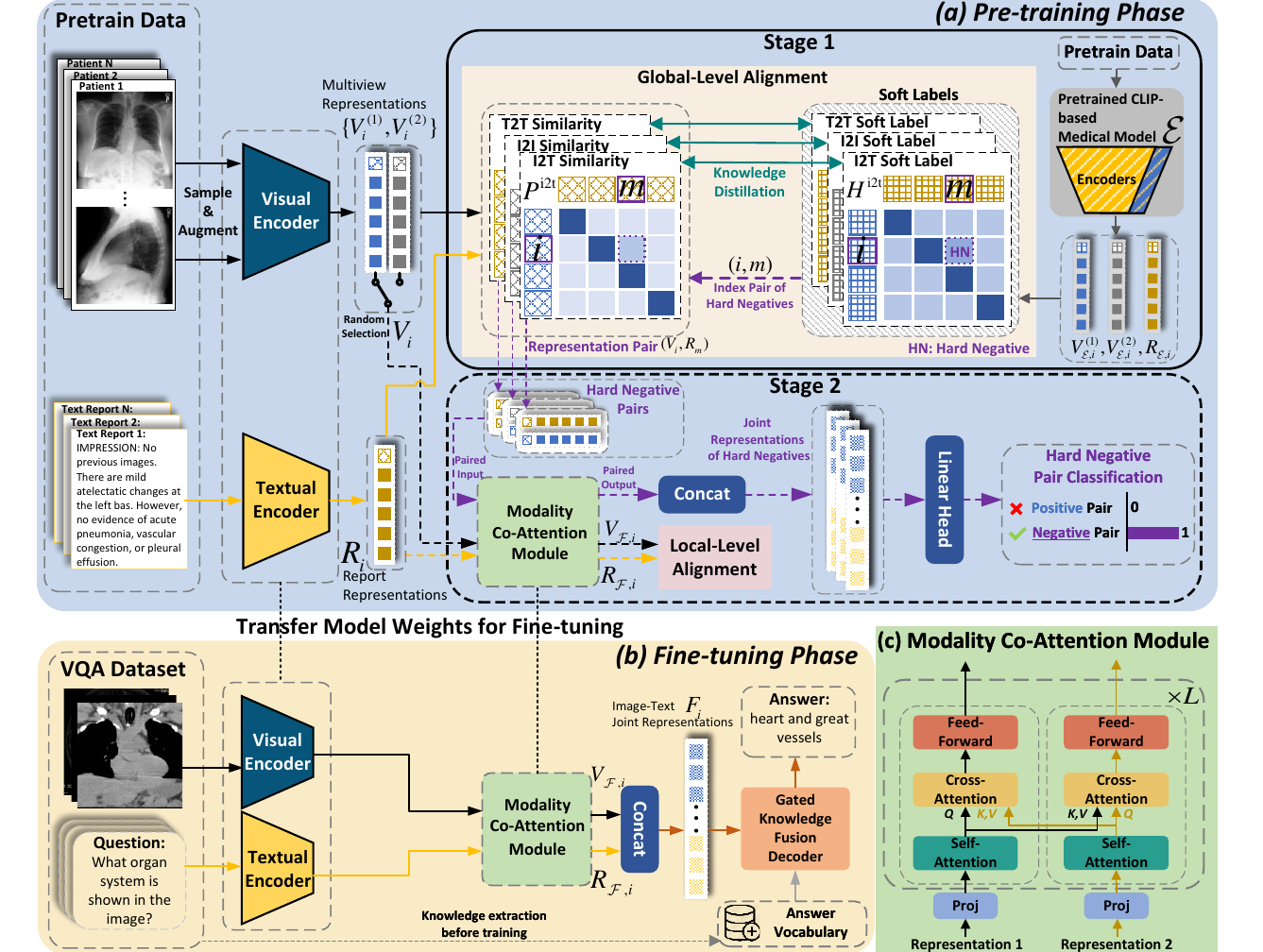}

   \caption{Architecture of AMiF: (a) Pre-Training Phase, (b) Fine-Tuning Phase, and (c) Modality Co-Attention Module.}
   \label{fig:archi}
\end{figure*}

\subsection{Overview}
\label{overview method}
As illustrated in \cref{fig:archi}, we divide the training process of AMiF into two phases: (a) pre-training and (b) fine-tuning.

In pre-training phase, we use the medical pre-training dataset $\{(I_i,W_i)\}_{i=1}^N$, where $N$ is the dataset sample number and $I_i=\{I_i^{(1)},\ldots,I_i^{(v_i)}\}$ is one imaging study case corresponding to one patient. We randomly sample multiple views from $I_i$, which corresponds to a single report $W_i$. If the case $I_i$ does not have multiple views (\ie, $v_i=1$), we augment the identical view multiple times. These multi-view images are then encoded by a visual encoder, resulting in multi-view representations $V_i^{(1)}, V_i^{(2)}$ (the number of multi-view is chosen as 2 in this paper), while the corresponding report is encoded as report representation $R_i$. We split the pre-training into two stages: \textbf{(1)} Global-Level Alignment with soft labels, and \textbf{(2)} Tokens are further matched between $V_i$ and $R_i$, where $V_i$ is a random selection between $V_i^{(1)}$ and $V_i^{(2)}$.

In the fine-tuning phase, our AMiF utilizes the initialized weights (\ie, visual encoder, textual encoder, and Modality Co-Attention Module) from the pre-training phase, and is then adapted to the Med-VQA task via a Gated Knowledge Fusion Decoder, which takes answer vocabulary from the specific Med-VQA dataset as the prior knowledge and generates answers via an auto-regressive decoder.

\subsection{Unified Modality Alignment}
\label{modality alignment}

\subsubsection{Global-Level Alignment}
\label{global level alignment}
Given the multi-view representations $V_i^{(1)}, V_i^{(2)}$ and text representation $R_i$ extracted by the visual and textual encoders, their \texttt{[CLS]} tokens are denoted as $\bs{v}_i^{(1)},\bs{v}_i^{(2)},\bs{r}_i \in \mathbb{R}^d$ respectively ($d$ is the token embedding dimension), which are used for the following inter-modality alignment (image-text) and intra-modality alignment (split as multi-view alignment and text-text alignment).

\paragraph{Inter-Modality (Image-Text)} 
First, we project the image and text embeddings $\bs{v}_i$ and $\bs{r}_i$ via a projection function $\mr{proj}(\cdot)$, where $\bs{v}_i\in \{\bs{v}_i^{(1)}, \bs{v}_i^{(2)}\}$. Then, we compute the cosine similarity between the $i$-th image sample and the $j$-th text sample as:
\begin{equation}
    s^{\mr{i2t}}_{ij}=\mr{proj}(\bs{v}_i)^\top\mr{proj}(\bs{r}_j).
    \label{cos sim-i2t-1}
\end{equation}
The probability of aligning (or matching) image i with text report j is computed as:
\begin{equation}
    P^{\mr{i2t}}_{ij}=\frac{\mr{exp}( s^{\mr{i2t}}_{ij}/\tau_1)}{\sum_{k=1}^N\mr{exp}( s_{ik}^{\mr{i2t}}/\tau_1)},
    \label{cos sim-i2t-2}
\end{equation}
where $\tau_1$ is a temperature parameter. The similar computation is applied to $P^{\mr{t2i}}_{ij}$.

\paragraph{Intra-Modality (Multi-View \& Text-Text)}
For the multi-view images, the probability of aligning image $\bs{v}^{(1)}_i$ and $\bs{v}^{(2)}_i$ is computed as:
\begin{equation}
\begin{aligned}
    s^{\mr{i1,i2}}_{ij}&=\mr{proj}(\bs{v}^{(1)}_i)^\top\mr{proj}(\bs{v}^{(2)}_j),
    \\
    P^{\mr{i1,i2}}_{ij}&=\frac{\mr{exp}( s^{\mr{i1,i2}}_{ij}/\tau_2)}{\sum_{k=1}^N\mr{exp}( s_{ik}^{\mr{i1,i2}}/\tau_2)}.
    \label{cos-sim-i2i}
\end{aligned}
\end{equation}
Similarly we obtain $P^{\mr{i2,i1}}_{ij}$. For text-text alignment, the probability of aligning report $i$ and report $j$ is computed as:
\begin{equation}
\begin{aligned}
    s^{\mr{t2t}}_{ij}&=\mr{proj}(\bs{r}_i)^\top\mr{proj}(\bs{r}_j),
    \\
    P^{\mr{t2t}}_{ij}&=\frac{\mr{exp}( s^{\mr{t2t}}_{ij}/\tau_3)}{\sum_{k=1}^N\mr{exp}( s_{ik}^{\mr{t2t}}/\tau_3)}.
    \label{cos-sim-t2t}
\end{aligned}
\end{equation}
The $\tau_2$ and $\tau_3$ are temperature parameters. 

\paragraph{Soft Label Generation}
We leverage a CLIP-based model \cite{zhang2023biomedclip} that is pre-trained on medical data to generate soft labels for the inter-modality and intra-modality alignment. Denoting the multi-view image and text representations of the $i$th sample from the pre-trained CLIP-based model $\mathcal{E}$ as $V^{(1)}_{\mathcal{E},i}, V^{(2)}_{\mathcal{E},i}$ and $R_{\mathcal{E},i}$, we use the embeddings of their \texttt{[CLS]} tokens as the global-level representations, yielding $\bs{v}^{(1)}_{\mathcal{E},i}, \bs{v}^{(2)}_{\mathcal{E},i}, \bs{r}_{\mathcal{E},i} \in \mathbb{R}^{d}$. Afterward, following the same routines as \cref{cos sim-i2t-2,cos-sim-i2i,cos-sim-t2t}, we compute the softmax cosine similarities $Q^{\mr{i2t}}_{ij},Q^{\mr{t2i}}_{ij},Q^{\mr{i1,i2}}_{ij},Q^{\mr{i2,i1}}_{ij},Q^{\mr{t2t}}_{ij}\in \mathbb{R}$ based on $\bs{v}^{(1)}_{\mathcal{E},i}, \bs{v}^{(2)}_{\mathcal{E},i}$ and $\bs{r}_{\mathcal{E},i}$. To add flexibility to the influence of positive image-text pairs and avoid classifying hard negatives as positive, we augment the soft labels by adding a flexibility weight $\lambda$ for the positive pairs. Taking the soft label for inter-modality $Q^{\mr{i2t}}_{ij}$ as an example, we have the soft label computed as:
\begin{equation}
\begin{aligned}
    H^{\mr{i2t}}_{ij}=Q^{\mr{i2t}}_{ij} + \lambda \cdot \mathbf{1}_{\{i=j\}},
\end{aligned}
\label{soft-label-modified}
\end{equation}
The same augmentation applies to all other soft labels, generating $H^{\mr{t2i}}_{ij}, H^{\mr{i1,i2}}_{ij}, H^{\mr{i2,i1}}_{ij}$ and $H^{\mr{t2t}}_{ij}$.   

\paragraph{Alignment with Soft Labels}
In contrastive learning, hard negatives are defined as negative image-text pairs with high similarities mistakenly. Since the one-hot label focuses on the positive image-text pair within the batch, the hard negatives are often overlooked. Instead of using one-hot labels and InfoNCE loss \cite{loss-infonce} as CLIP-based methods, we use soft labels for both inter-modality and intra-modality alignment. Specifically, We employ the knowledge distillation \cite{kd}, comparing the similarity gained in \cref{cos sim-i2t-2,cos-sim-i2i,cos-sim-t2t} with the corresponding soft labels from the pre-trained CLIP-based model $H^{\mr{i2t}}_{ij}, H^{\mr{t2i}}_{ij}, H^{\mr{i1,i2}}_{ij}, H^{\mr{i2,i1}}_{ij}, H^{\mr{t2t}}_{ij}\in \mathbb{R}$. The knowledge distillation is achieved by minimizing the KL-divergence loss, computed as:
\begin{equation}
\resizebox{0.9\hsize}{!}{$
\begin{split}
    \mathcal{L}_{\mr{inter}}&=\frac12\left(\mathcal{L}_{\mr{inter}}^{\mr{i2t}}+\mathcal{L}_{\mr{inter}}^{\mr{t2i}}\right)
    \\
    &=\frac{1}{2N} \sum_{i=1}^N\left(\mr{KL}(H^{\mr{i2t}}_{i} \Vert P^{\mr{i2t}}_{i})+\mr{KL}(H^{\mr{t2i}}_{i}\Vert P^{\mr{t2i}}_{i})\right),
\end{split}$}
    \label{kl div loss inter modality}
\end{equation}
where $P^{\mr{i2t}}_{i}=\left[P^{\mr{i2t}}_{i1}, \ldots,P^{\mr{i2t}}_{ij},\ldots,  P^{\mr{i2t}}_{iN}\right]$ denotes the probability distribution of aligning image $i$ with all $N$ possible reports, and the similar representation applies for $P^{\mr{t2i}}_{i}, H^{\mr{i2t}}_{i} $ and $H^{\mr{t2i}}_{i}$. The loss for intra-modality alignment is:
\begin{equation}\resizebox{0.88\hsize}{!}{$
\begin{split}
    \mathcal{L}_{\mr{intra}}=&\frac12(\mathcal{L}_{\mr{intra}}^{\mr{i1,i2}}+\mathcal{L}_{\mr{intra}}^{\mr{i2,i1}})+\mathcal{L}_{\mr{intra}}^{\mr{t2t}}
    \\
    =&\frac{1}{2N} \sum_{i=1}^N(\mr{KL}(H^{\mr{i1,i2}}_{i} \Vert P^{\mr{i1,i2}}_{i})+\mr{KL}(H^{\mr{i2,i1}}_{i}\Vert P^{\mr{i2,i1}}_{i}))
    \\
    &+\frac{1}{N}\sum_{i=1}^N\mr{KL}(H^{\mr{t2t}}_{i}\Vert P^{\mr{t2t}}_{i}).
\end{split}$}
\label{kl div loss intra modality}
\end{equation}
For the first stage of pre-training, we conduct the global-level alignment with the following total loss: 
\begin{equation}
\mathcal{L}_{\mr{global}}=\mathcal{L}_{\mr{inter}}+\mathcal{L}_{\mr{intra}}.
\end{equation}

\subsubsection{Local-Level Alignment}
In the second stage of pre-training, we first pass the multi-modal representations $V_i$ and $R_i$ through a Modality Co-Attention Module (\cref{fig:archi} (c)) that fuse each other's information, yielding the mutually-aware feature representations $V_{\mathcal{F},i}$ and $R_{\mathcal{F},i}$. Next, we align the $p$ token embeddings of the image representation $V_{\mathcal{F},i}=\{\bs{v}_{\mathcal{F},i}^1, \bs{v}_{\mathcal{F},i}^2, \ldots, \bs{v}_{\mathcal{F},i}^p\}$ with the $t$ token embeddings of the report representation $R_{\mathcal{F},i}=\{\bs{r}_{\mathcal{F},i}^1, \bs{r}_{\mathcal{F},i}^2, \ldots, \bs{r}_{\mathcal{F},i}^t\}$.

Our objective is to compute an optimal transport plan $\mathbf{T}\in\mathbb{R}^{p\times t}$ that minimizes the following loss function:
\begin{equation}
\resizebox{0.88\hsize}{!}{$
\begin{aligned}
\mathcal{L}_{\mr{local}} = \frac{1}{N} \sum_{i=1}^N \min\limits_{\mathbf{T}} \sum_{j=1}^{p} \sum_{k=1}^{t} \mathbf{T}_{jk} \cdot \mr{C} \left( \bs{v}_{\mathcal{F},i}^j, \bs{r}_{\mathcal{F},i}^k \right),
\end{aligned}$}
\label{ipot-1}
\end{equation}
where $\mr{C}(\cdot)$ is the cost function between two embeddings:
\begin{equation}
\mr{C} \left( \bs{v}_{\mathcal{F},i}^j, \bs{r}_{\mathcal{F},i}^k \right) = 1 - (\bs{v}_{\mathcal{F},i}^j)^\top \bs{r}_{\mathcal{F},i}^k.
\label{ipot-cost-function}
\end{equation}

Since minimizing the transport plan $\mathbf{T}$ directly is intractable, we use the inexact proximal point method (IPOT, \cite{boyd2004convex, uniter, ipot-algorithm}) to approximate an optimal transport plan $\mathbf{T}^*$. The resulting optimal plan $\mathbf{T}^*$ captures how each image patch token is associated with a word token, based on the elements of the matrix.

\subsection{Hard Negative Mining}
\label{hard negative mining}
The soft labels used for global-level alignment might compute high similarities to both hard negative pairs and positive pairs, which may lead to difficulties in distinguishing hard negative pairs from positive ones. We pair the representations of hard negatives from both inter-modality and intra-modality alignments, passing them through the Modality Co-Attention Module to generate mutually-aware representation pairs and enforce the classification on these representation pairs to be negative pairs, as illustrated in the lower-right of \cref{fig:archi} (a).

Specifically, take the $i$-th sample for example, which consists of an image-text pair $(V_i,R_i)$ used for its multi-modality alignment. The distribution of I2T (image-text) soft labels for $V_i$ is $H^{\mr{i2t}}_{i} = \left[ H^{\mr{i2t}}_{i1}, \ldots, H^{\mr{i2t}}_{ij}, \ldots, H^{\mr{i2t}}_{iN} \right]$. We sample a hard negative pair $(V_i,R_m)$ from $H^{\mr{i2t}}_{i}$. The sampling rule is that a negative pair with higher similarity has a higher chance of being chosen. Based on the index pair $(i,m)$ of the hard negative pair, where $m \neq i$, we pair $(V_i,R_m)$ together, passing them to the Modality Co-Attention Module, resulting in a mutually-aware representation pair $(\bs{V}_{\mathcal{F},i}, \bs{R}_{\mathcal{F},m})$. Similarly, for intra-modality alignment, we consider hard negative pairs within multi-view images and texts. For multi-view images, suppose that $V_i$ of the $i$-th sample and $V_n$ of the $n$-th sample $(n\neq i)$ form a hard negative pair. For texts, we have $R_i$ and $R_k$ forming a hard negative pair similarly, where $i \neq k$. Therefore, for the $i$-th sample, we build a hard negative set:
\begin{equation}\resizebox{0.88\hsize}{!}{$\begin{split}
    \mathcal{S}_i^{\mr{HN}}=\{(\bs{V}_{\mathcal{F},i}, \bs{R}_{\mathcal{F},m}),(\bs{V}_{\mathcal{F},i}, \bs{V}_{\mathcal{F},n}),(\bs{R}_{\mathcal{F},i}, \bs{R}_{\mathcal{F},k})\}.
\end{split}$}
\label{hard negative set}
\end{equation}

Next, we formulate a classification problem to enforce the discrimination between hard negative pairs and positive pairs. Given representations of any hard negative pair $(\bs{a},\bs{b})\in \mathcal{S}_i^{\mr{HN}}$, we first concatenate ($\mr{Concat}$) these two representations, forming a joint representation. Then, a linear classification head ($\mr{LH}$) is applied to this joint representation, which outputs a prediction logits:
\begin{equation}
    \bs{p}_{(\bs{a},\bs{b})}=\mr{LH}(\mr{Concat}(\bs{a},\bs{b})).
    \label{classification logits}
\end{equation}
A cross-entropy ($\mr{CE}$) loss is applied on classifying hard negative pairs using \cref{classification logits}:
\begin{equation}
\begin{aligned}
       \mathcal{L}_{\mr{HN}}= \frac1N\sum_{i=1}^N\sum_{(\bs{a},\bs{b})\in \mathcal{S}_i^{\mr{HN}}} \mr{CE}(\bs{p}_{(\bs{a},\bs{b})};y_{(\bs{a},\bs{b})}),
\end{aligned}
\label{hard negative ce}
\end{equation}
where $y_{(\bs{a},\bs{b})}$ is a one-hot label (the hard negative pair is assigned with label $1$ while the positive pair with label $0$).

\subsection{Gated Knowledge Fusion Decoder}
To fine-tune the model pre-trained by the alignment and hard negative mining for specific downstream tasks such as Med-VQA (\cref{fig:archi} (b)), we design a Gated Knowledge Fusion Decoder (\cref{fig: gca}), which consists of (1) a Gated Cross-Attention Module to select related knowledge from the generic prior knowledge and fuse it into the domain-specific downstream task; and (2) an auto-regressive decoder to generate \textit{open}-form answers.

\subsubsection{Gated Cross-Attention Module}
\label{Gated Cross-Attention}

\begin{figure}[t]
  \centering
   \includegraphics[width=1\linewidth]{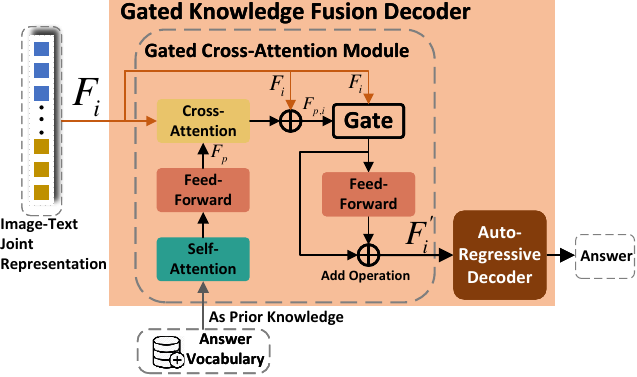}

   \caption{Architecture of Gated Knowledge Fusion Decoder, which consists of a Gated Cross-Attention Module and an Auto-Regressive Decoder.}
   \vspace{-.2in}
   \label{fig: gca}

\end{figure}

Generic prior knowledge usually contains comprehensive information covering a wide range of medical topics, so irrelevant information is involved when applied to a specific-field Med-VQA task. Thus, we propose a Gated Cross-Attention Module (\cref{fig: gca}), which fuses the prior knowledge into the input image-question representation and uses a gate operation to select relevant information from the knowledge-fused representation.

Given an image-text joint representation $F_{i} \in \mathbb{R}^{(p+t) \times d}$, where $p,t$ are the token numbers for image and text representations. We input it into the Gated Cross-Attention Module. The generic prior knowledge is represented by an answer vocabulary extracted from the Med-VQA dataset for fine-tuning, which is passed through a self-attention and a feed-forward sub-module, yielding $F_p \in \mathbb{R}^{M \times d}$, where $M$ is the total token length of the knowledge. We use cross-attention for fusing the prior knowledge $F_p$ and joint representation $F_i$. Specifically, the knowledge-fused representation is computed as:
\begin{equation}
\begin{aligned}
    F_{p,i} = \mr{CA}(F_i, F_p) + F_i,
\end{aligned}
\label{gca-1}
\end{equation}
where $\mr{CA}(\cdot)$ represents the cross-attention mechanism, with the first argument as the query (Q), and the second as the key (K) and value (V). 

Next, to select related information from prior knowledge for accurate answer prediction, we have:
\begin{equation}
\begin{aligned}
    G_{p,i} = \mr{Sigmoid}(\mr{Projection}(F_{p,i})),
\end{aligned}
\label{gca-2}
\end{equation}
where $\mr{Sigmoid}(\cdot)$ is an activation function and $\mr{Projection}(\cdot)$ is a mapping function in the $d$-dimensional space. The resulting $G_{p,i}\in \mathbb{R}^{(p+t) \times d}$ is a weight matrix with all values between 0 and 1, indicating the portion of information selected from knowledge fused representation $F_{p,i}$. To balance information from the knowledge-fused representation $F_{p,i}$ and the image-text joint representation $F_i$, we have the gated operation in \cref{fig: gca} formulated as:
\begin{equation}
\resizebox{0.88\hsize}{!}{$\begin{split}
    \mr{Gate}(F_{p,i}, F_i) = G_{p,i} \odot F_{p,i} + (1 - G_{p,i}) \odot F_i,
\end{split}$}
\label{gca-3}
\end{equation}
where $\odot$ is the element-wise product. The representation $\mr{Gate}(F_{p,i}, F_i)$ is then passed through a feed-forward sub-module, yielding a final knowledge-fused image-question representation $F'_{i}$ for the answer generation below.

\subsubsection{Auto-Regressive Decoder}
Different from prior methods like \cite{m3ae,m2i2,arl} that predict answers by classifying from a fixed answer list, we utilize an auto-regressive decoder \cite{albef,cho2021unifying} to generate \textit{open}-form answers. Given the selected knowledge-fused image-question representation $F'_i$ from \cref{Gated Cross-Attention}, the decoder is trained to generate answers auto-regressively via a conditional language modeling loss.

%% file: sec/4_Experiment.tex
\section{Experiment}
\label{experiment}

\begin{table*}[t]
  \centering
  \renewcommand{\arraystretch}{1.13}
  \fontsize{8}{8}\selectfont
  \begin{adjustbox}{width=\textwidth}
  \begin{threeparttable}
    \begin{tabular}{c|cc|c|cc|c|cc|c|c}
    \Xhline{1.2pt}
    \multirow{2}{*}{Method}&
    \multicolumn{3}{c|}{VQA-RAD}&\multicolumn{3}{c|}{SLAKE}&\multicolumn{3}{c|}{PathVQA}& \multicolumn{1}{c}{VQA2019} \\
    \cline{2-11}
    &Open &Closed &Overall &Open &Closed &Overall &Open &Closed &Overall&Overall\\
   \hline
    MMQ \cite{mmq2021}& 53.70& 75.80& 67.00&- & -&-&13.40 &84.0& 48.80&-\\
    MEVF-BAN \cite{nguyen2019mevf}&49.20&77.20&66.10&77.80&79.80&78.60&-&-&-&77.86\\
    CPRD \cite{liu2021cprd}& 52.50& 77.90& 67.80& 79.50& 83.40& 81.10 & - & - & -& -\\
    MMBERT \cite{khare2021mmbert}& 63.10 &77.90 &72.00& -& -& - & - & - & -& 67.20\\
    ARL \cite{arl}& 65.10 & 85.96 & 77.50 & 79.70 & 89.30 & 84.00 & - & - & - &80.32\\
    M3AE \cite{m3ae}& 67.23 & 83.46 & 77.01 & 80.31 & 87.82 & 83.25 & - & - & - & 79.87\\
    PubMedCLIP \cite{pubmedclip}& 60.10 & 80.00 & 72.10 & 78.40 & 82.50 & 80.10 & - & - & -& -\\
    CPCR \cite{cpcr}& 60.50& 80.40 &72.50& 80.50& 84.10 &81.90&- &- & -&-\\
    M2I2 \cite{m2i2}& 61.80 & 81.60 & 73.70 & 74.70 & \textbf{91.10} & 81.20 & 36.30 & 88.00 & 62.20& -\\
    MUMC \cite{mumc}& \underline{71.50} & 84.20 & 79.20 & - & - & 84.90 & \underline{39.00} & \underline{90.40} & \underline{65.10} & - \\
    RAMM \cite{yuan2023ramm} & - &- &78.27& -& -& \underline{86.05} & - & - & -& \underline{82.13}\\
    PeFoMed \cite{pefomed}& 62.60 & \underline{87.10} & 77.40 & 77.80 & 88.70 & 82.10 & 35.70 & \textbf{91.30} & 63.60 &-\\
    LaPA \cite{lapa}& 68.72 & 86.40 & \underline{79.38} & 82.17 & 88.70 & 84.73 & - & - & - &81.60\\
    BiomedCLIP \cite{zhang2023biomedclip}& 67.60& 79.80& 75.20& \underline{82.50}& 89.70 &85.40&- &- & -& -\\
   \hline
    AMiF(Ours)&\textbf{72.00}&\textbf{87.25}&\textbf{80.49}&\textbf{84.84}&\underline{90.23}&\textbf{86.71}&\textbf{40.33}&90.18&\textbf{65.33}&\textbf{83.20}\\
    \Xhline{1.2pt}
    \end{tabular}
    \end{threeparttable}
    \end{adjustbox}
\caption{Accuracy comparison with existing methods on various VQA datasets, evaluated for \textit{Open} and \textit{Closed} sets and the \textit{Overall} performances. \textit{Open} represents a question set with \textit{open}-form answers while \textit{Closed} represents the question set with \textit{yes/no} as the answer. The best and second-best results are \textbf{bolded} and \underline{underlined} respectively.}
\label{table:sota-vqa}
\end{table*}

\subsection{Datasets}
\label{dataset}
\paragraph{Pre-training Datasets}
For pre-training AMiF, we use three medical datasets that are widely used for Med-VLP: MIMIC-CXR \cite{data-mimic} is the largest publicly available radiology dataset that contains 377110 X-ray chest images and 227827 corresponding reports; ROCO \cite{data-roco} is a radiology dataset that has over 81,000 radiology images from various imaging modalities and their corresponding captions; and MedICaT \cite{data-medicat} is a medical figure-caption dataset that contains over 217,000 image-caption pairs.

\paragraph{Multi-view Images}
For MIMIC-CXR, there are existing multi-view images for each radiology study, so we randomly sample two different views. For ROCO and MedICaT datasets, which lack the multi-view information for each sample, we apply strong random data augmentation \cite{simclr} for generating multi-views from one image. Ultimately we ensure all samples from the datasets have two multi-view images for intra-modality (image-image) alignment.

\paragraph{Med-VQA Datasets for Fine-tuning}
For fine-tuning the model for the Med-VQA task, we use VQA-RAD \cite{data-vqarad}, SLAKE \cite{data-slake}, VQA-2019 \cite{data-vqa2019} and PathVQA\cite{data-pathvqa}. VQA-RAD has 315 radiology images with 3064 question-answer pairs. SLAKE has 14,028 pairs of samples. VQA-2019 comprises 15,292 question-answer pairs. PathVQA contains 32,799 question-answer pairs. All datasets contain \textit{Open} set (open-form answers) and \textit{Closed} set (\textit{yes/no} answers), and we follow the official splits for all datasets.

\subsection{Implementation Details}
\label{implementation}
For the pre-training phase, we apply CLIP-ViT/B-16 \cite{clip} as the initialized visual encoder, and ClinicalBERT \cite{huang2019clinicalbert} as the initialized textual encoder and decoder. For the fine-tuning phase, the Modality Co-Attention Module uses $L=6$ layers. The temperature parameters $\tau_1,\tau_2,\tau_3$ are set to 0.07. The flexibility weight of the positive pair $\lambda$ is set to $0.01$. The total trainable model size is 326M, while the size of Gated Cross-Attention Module are 14.8M. 

Our pre-training phase contains 20 epochs for stage one, and 10 epochs for stage two, with 256 as batch size. The model uses AdamW optimizer with a learning rate of 4e-6. 

For more details about the implementation and training, please refer to the Supplement.

\begin{table*}[t]
    \centering
    \renewcommand{\arraystretch}{1.15}
    \fontsize{8}{8}\selectfont 
    \begin{adjustbox}{width=0.6\textwidth} 
        \begin{threeparttable}
            \begin{tabular}{c|ccccc|cccc|c}
                \Xhline{1.2pt}
                \textbf{\#} & G-inter & G-intra & Local & Soft Label & HN & RAD-VQA & SLAKE & Path-VQA & VQA-2019 & Avg \\
                \hline
                1 & \checkmark &  &  &  &  & 76.68 & 82.4 & 61.23 & 79.06 & 74.84 \\
                2 & \checkmark & * &  &  &  & 77.06 & 82.47 & 61.92 & 79.70 & 75.28 \\
                3 & \checkmark & \checkmark &  &  &  & 77.92 & 82.64 & 63.78 & 79.88 & 76.06 \\
                4 & \checkmark & \checkmark & \checkmark &  &  & 78.44 & 83.83 & 64.66 & 80.92 & 76.96 \\
                5 & \checkmark &  & \checkmark & \checkmark &  & 79.04 & 85.98 & 65.24 & 81.34 & 77.90 \\
                6 & \checkmark &  & \checkmark & \checkmark & \checkmark & 79.77 & \underline{86.20} & \underline{65.30} & 81.83 & 78.28 \\
                7 & \checkmark & \checkmark & \checkmark & \checkmark &  & \underline{79.82} & 86.07 & 65.21 & \underline{82.21} & \underline{78.33} \\
                8 & \checkmark & \checkmark & \checkmark & \checkmark & \checkmark & \textbf{80.49} & \textbf{86.71} & \textbf{65.33} & \textbf{83.20} & \textbf{78.93} \\
                \Xhline{1.2pt}
            \end{tabular}
            \caption{Ablation study for pre-training using different components (\checkmark). G-inter and G-intra denote global inter- and intra-modality alignments, respectively. For intra-modality, * denotes only text-text alignment, while \checkmark denotes both multi-view and text-text alignments. Local denotes local-level alignment. Soft Label denotes soft label for global-level alignment and HN denotes hard negative mining.}
        \label{Ablative pre-training}
        \end{threeparttable}
    \end{adjustbox}
    \hfill
    \renewcommand{\arraystretch}{1.3}
    \fontsize{8}{8}\selectfont
    \begin{adjustbox}{width=0.39\textwidth} 
        \begin{threeparttable}
            \begin{tabular}{c|cc|cccc|c}
            \Xhline{1.2pt}
                \textbf{\#} & $\mathcal{L}_{\mr{local}}$ & $\mathcal{L}_{\mr{HN}}$ & RAD-VQA & SLAKE & Path-VQA & VQA-2019 & Avg \\
                \hline
                1 & 1.0 & 1.0 & \textbf{80.49} & \textbf{86.71} & \underline{65.33} & \textbf{83.20} & \textbf{78.93} \\
                2 & 1.0 & 0.2 & 78.32 & \underline{84.92} & 63.30 & 82.15 & 77.17 \\
                3 & 0.2 & 1.0 & \underline{79.57} & 84.82 & 64.93 & 82.53 & \underline{77.96} \\
                4 & 1.0 & 3.0 & 79.16 & 85.17 & 64.46 & 82.26 & 77.76 \\
                5 & 3.0 & 1.0 & 78.61 & 84.59 & \textbf{65.42} & \underline{83.14} & 77.94 \\
            \Xhline{1.2pt}
            \end{tabular}
             \caption{Ablation study for pertaining with different weight combinations of Hard Negative Mining Loss (HN) and Local-Level Alignment Loss (local).}
             \label{weight ablation local hard negative}
        \end{threeparttable}
    \end{adjustbox}
    \vspace{-.1in}
\end{table*}

\subsection{Comparison with the State-of-the-Art}
\label{comparison sota}
As shown in \cref{table:sota-vqa}, our method outperforms other methods across all Med-VQA datasets for \textit{Overall} accuracy, and for the \textit{Open} type questions, which is more challenging than \textit{Closed} type, we also achieve the best performance. Notably, compared with BiomedCLIP \cite{zhang2023biomedclip}, which is the pre-trained model we utilize to generate soft labels, our method surpasses it on VQA-RAD and SLAKE datasets by 5.29\% and 1.31\% separately. For the PathVQA dataset, which is the largest one with challenging open-form question-answer pairs, we outperform the second best by 1.33\%. These results show that our method has a better understanding ability on the Med-VQA tasks, with room for improvements. 
\begin{table}[h!]
    \centering
    \renewcommand{\arraystretch}{1.1}
    \fontsize{7}{7}\selectfont
    \resizebox{\linewidth}{!}{
    \begin{threeparttable}
            \begin{tabular}{c|cc|cccc|c}
                \Xhline{1.2pt}
                \textbf{\#} & Pre-training & Knowledge & RAD-VQA & SLAKE & Path-VQA & VQA-2019 & Avg \\
                \hline
                1 & - & - & 75.64 & 81.09 & 60.92 & 78.14 & 73.95 \\
                2 & CLIP-based & - & 77.65 & 82.92 & 61.30 & 79.92 & 75.45 \\
                3 & Ours & - & 79.03 & 84.32 & 63.53 & 81.03 & 76.98 \\
                4 & Ours & \textbf{Q} & 79.55 & 85.04 & 63.93 & \underline{82.13} & 77.36 \\
                5 & Ours & \textbf{A} & \underline{80.26} & \underline{85.37} & \underline{63.96} & 81.26 & \underline{77.71} \\
                6 & Ours & \textbf{A}\&\textbf{G} & \textbf{80.49} & \textbf{86.71} & \textbf{65.33} & \textbf{83.20} & \textbf{78.93} \\
                \Xhline{1.2pt}
            \end{tabular}
             \caption{Ablation study for fine-tuning on Med-VQA task. \textbf{Q} and \textbf{A} denotes introducing question and answer vocabulary as prior Knowledge, respectively. \textbf{G} denotes using the \textbf{G}ated operation in the Gated Cross-Attention Module.}
        \label{Ablative fine-tuning VQA}
        \end{threeparttable}
    }
\end{table}
\subsection{Ablation Study}
\label{ablation study}
\paragraph{Ablation for Unified Modality Alignment}
To explore the impact of different components for pre-training, we conduct systematic ablative experiments for unified modality alignments in \cref{Ablative pre-training}. Notably, our baseline (\#1) only uses global inter-modality alignment. To prove the significance of \ul{intra-modality alignment} to our AMiF, we conduct ablations \#3, \#7, and \#8, which show improvements in average accuracy compared with \#1, \#5, and \#6, respectively. When applying intra-modality alignment, we also evaluate the effectiveness of \ul{multi-view alignment} via \#3, which elevates the average accuracy by 0.78\% compared with only text-text alignment (\#2). To demonstrate that \ul{local-level alignment} improves the model's ability, we compare \#4 with \# 3, which shows a 0.90\% increase in accuracy. Moreover, We also evaluate the impact of different \ul{weight combination in stage 2 of pre-training} (\ie, local-level alignment loss and hard-negative mining loss), as shown in \cref{weight ablation local hard negative}. Based on the ablative results we choose to balance those two losses.

\begin{figure}[t]
  \centering
   \includegraphics[width=1\linewidth]{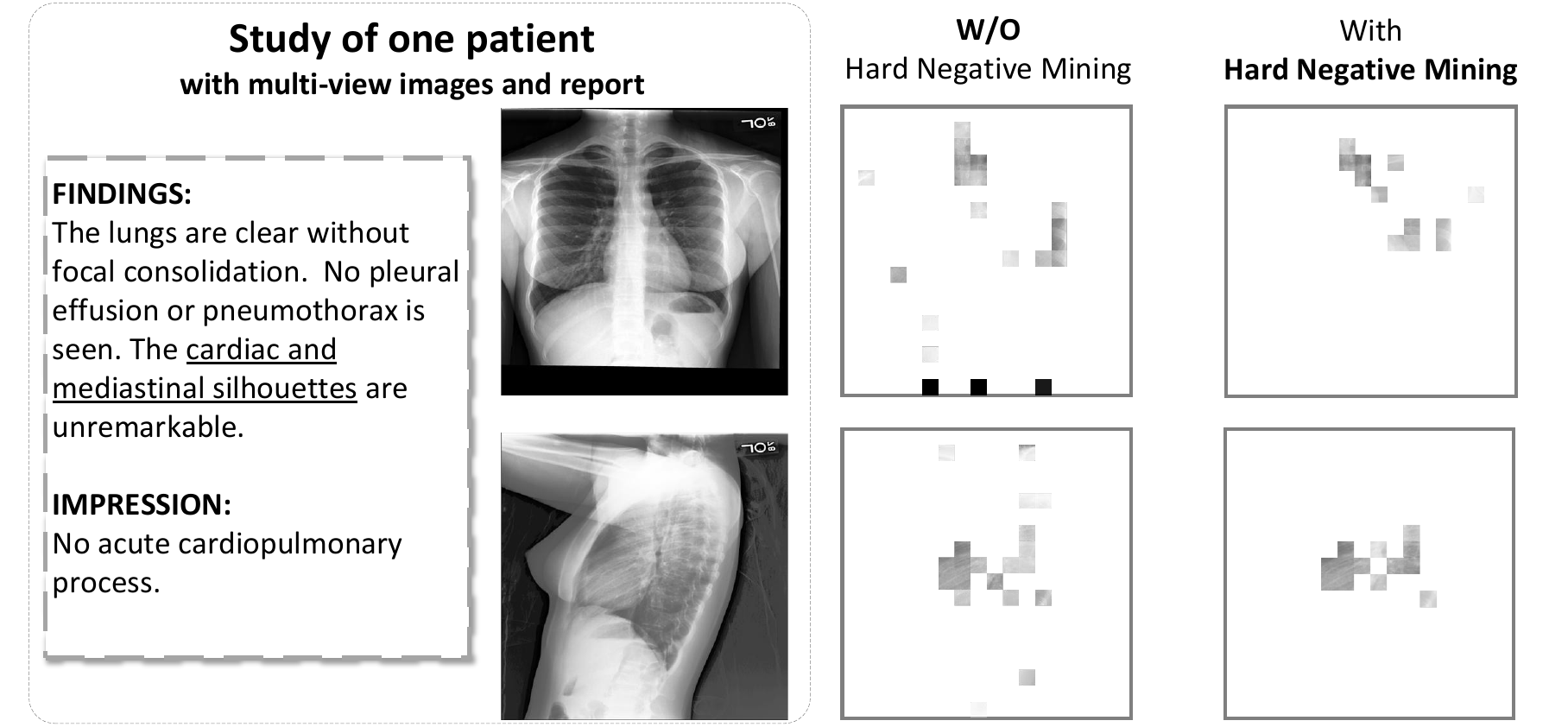}
   \caption{Visualization of the local-level alignment related to \underline{underlined} words for the ablative study of hard negative mining.}
   \vspace{-.1in}
   \label{fig: hard negative mining local alignment}
\end{figure}

\paragraph{Ablation for Hard Negative Mining}
To show that our approach of hard negative mining benefits model performance on Med-VQA tasks, we first investigate the necessity of using \ul{soft labels for global-level alignment}. As shown in \cref{Ablative pre-training}, the comparison between \#7 and \#4 indicates that using soft labels boosts the average model performance by 1.37\%. We also show that \ul{soft labels for global intra-modality alignment} facilitates the model's representation ability by comparing \#7 with \#5. Moreover, to prove the significance of \ul{hard negative pair discrimination}, we compare \#6 with \#5 and \#8 with \#7. Besides, we also conduct a qualitative ablation study for hard negative mining as shown in \cref{fig: hard negative mining local alignment}, comparing ablations \#8 and \#4. After hard negative mining, the patch tokens related to the underlined words are more concentrated on regions with possible semantic information, showing that our mining method leads to a more robust modality co-attention module, which yields more interpretable results for local-level alignment.

\begin{figure}[t]
  \centering
   \includegraphics[width=1\linewidth]{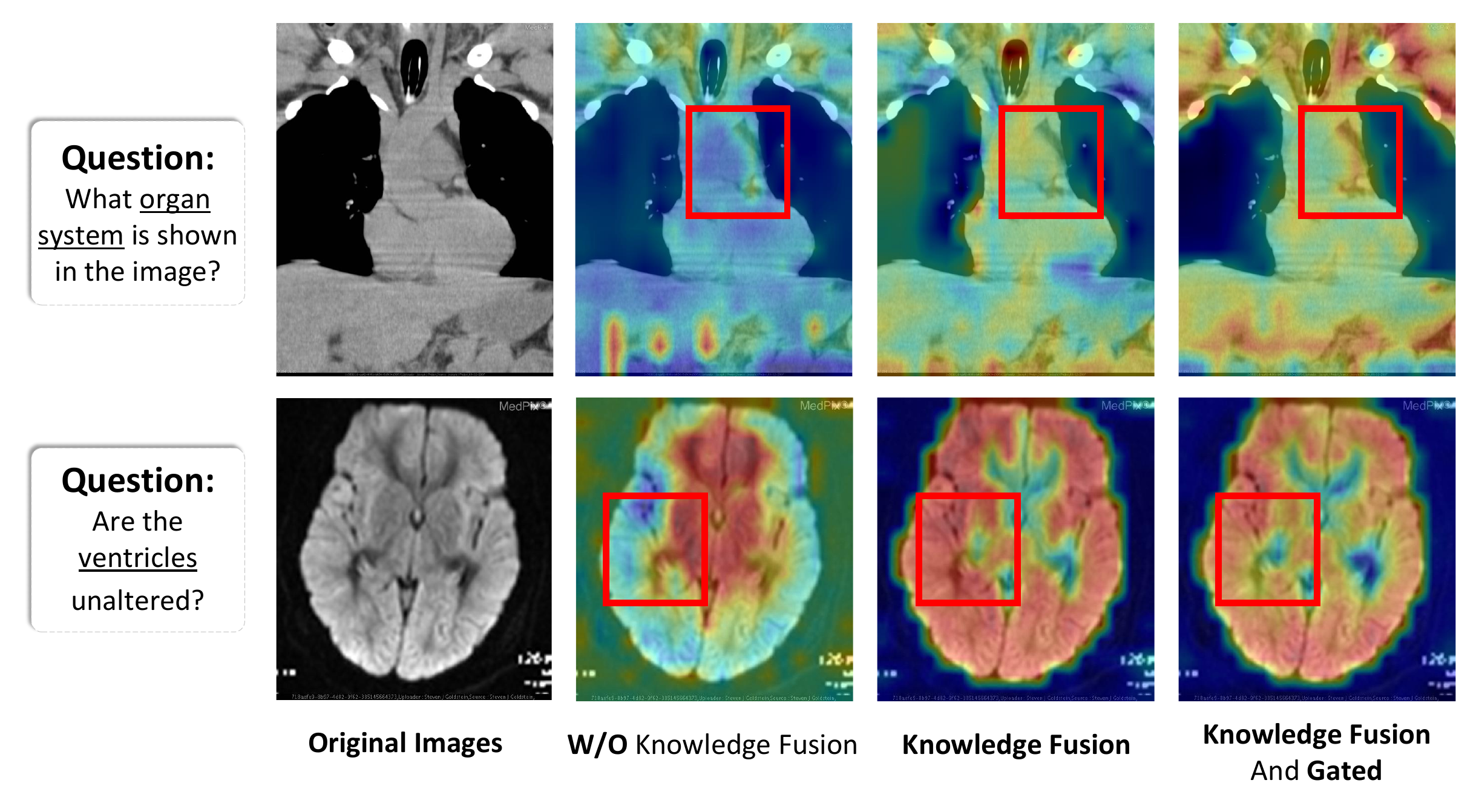}
   \caption{Attention map related to \underline{underlined} words for the ablative study of selective knowledge fusion.}

   \label{fig:ablation attention map}
\end{figure}

\paragraph{Ablation for Fine-tuning Med-VQA}
In \cref{Ablative fine-tuning VQA}, we show that compared with training our model from scratch (\#1), our \ul{pre-training} method (\#3) greatly improves the average accuracy by 3.03\%. The comparison between \#3 and \#2 shows that our pre-training method outperforms the CLIP-based model \cite{zhang2023biomedclip} used for generating soft labels. Besides the pre-training method, we study how \ul{selective knowledge fusion} facilitates Med-VQA tasks. Comparing \#4 with \#3, we find that fusing \ul{question vocabulary} to image-question representation improves the model performance for question-answering by 0.38\%. When we fuse the \ul{answer vocabulary} instead (\#5), we find the average performance is raised by 0.35\%. These results are further enhanced when we apply the gated operation in our \ul{Gated Cross-Attention Module} (\#6), improving the average accuracy by 1.22\% based on \#5, which is the best result we currently have. Additionally, we visualize the attention map in \cref{fig:ablation attention map}. From regions denoted with red boxes, we see that the gated operation selects relevant information and filters out unnecessary ones.

%% file: sec/5_Conclusion.tex
\section{Conclusion}
\label{sec:conclusion}

In this paper, we introduced \textbf{AMiF} to address the challenges in Medical Visual Question Answering (Med-VQA). Specifically, we propose solutions for unified modality \textbf{A}lignment, hard negative \textbf{Mi}ning, and selective knowledge \textbf{F}usion. We unify heterogeneous modality alignments across multiple levels, modalities, views, and stages. Our hard negative mining method employs soft labels for both inter-modality and intra-modality alignments and explores hard negative pairs to distinguish them from positive ones via enforced discrimination. For the Med-VQA task, we introduce a Gated Cross-Attention Module that integrates the answer vocabulary as prior knowledge and selects relevant information from it. Our experiments demonstrate the effectiveness of our approach on Med-VQA tasks, compared with the previous state-of-the-art.

%% file: sec/X_suppl.tex
\clearpage
\setcounter{page}{1}
\maketitlesupplementary

\section{Implementation Details}
For the pre-training phase, we apply CLIP-ViT/B-16 \cite{clip} as the initialized visual encoder, and ClinicalBERT \cite{huang2019clinicalbert} as the initialized textual encoder and auto-regressive decoder. All images are resized to 288×288. The dimension of embeddings $d$ is set to 768. The Modality Co-Attention Module uses $L=6$ layers. The total trainable model size is 326M. The temperature parameters $\tau_1,\tau_2,\tau_3$ are set to 0.07. The flexibility weight of the positive pair is cross-validated among different value choices and set $\lambda=0.01$. Our pre-training phase contains 20 epochs for stage one, and 10 epochs for stage two, with 256 as batch size. The model uses AdamW optimizer with a learning rate of 4e-6.

For the fine-tuning phase, the parameters for the Gated Cross-Attention Module are 14.8M, while 122M for the auto-regressive decoder. We fine-tune the model for 60 epochs with a learning rate of 2e-5 and a batch size of 64.

Our computing resources are two NVIDIA A100 GPUs, and the training uses 16-mixed precision. The pre-training costs roughly 24 hours, while fine-tuning time is around 8 hours. The inference time for one image-question pair is around 0.27$s$. During inference, to fairly compare with other multi-label classification-based methods, we apply the test set answers as the candidate answers, compare the generated open-form answer from our auto-regressive decoder with these candidates, and choose the one with the lowest language modeling loss.

\section{Additive Ablation Studies}
\label{sec:add ablative}
Besides the ablation study we conduct in \cref{ablation study}, we also evaluate the impacts of different methods we use to generate the soft labels, and how the $\lambda$ (\cref{soft-label-modified}) will affect the learning of these soft labels.

\begin{table}[h]
\renewcommand{\arraystretch}{1.3}
    \fontsize{8}{8}\selectfont
    \begin{adjustbox}{width=1\linewidth} 
        \begin{threeparttable}
            \begin{tabular}{c|c|cccc|c}
                \Xhline{1.2pt}
                \textbf{\#} & Soft Label Generation Method & RAD-VQA & SLAKE & Path-VQA & VQA-2019 & Avg \\
                \hline
                1 & - &  75.64 & 81.09 & 60.92 & 78.14 & 73.95 \\
                2 & CLIP\cite{clip}  & 76.35 & 81.32 & 61.38 & 78.42 & 74.37 \\
                3 & MedCLIP \cite{medclip}  & 78.80 & 84.82 & 62.83 & 81.03 & 76.87 \\
                4 & PubMedCLIP\cite{pubmedclip}  & 80.02 & 85.32 & 63.12 & 80.93 & 77.35 \\
                5 & PMC-CLIP\cite{pmc-clip}  & 79.63 & 84.98 & 63.01 & 80.44 & 77.02 \\
                6 & BioMedCLIP \cite{zhang2023biomedclip}  & 80.26 & 85.37 & 63.96 & 81.26 & \textbf{77.71} \\
                \Xhline{1.2pt}
            \end{tabular}

        \end{threeparttable}
    \end{adjustbox}
    \caption{Ablation study on different choices of soft label generation methods for Med-VQA tasks.}
    \label{tab: ablative soft label generation}
\end{table}

\subsection{Impacts of Different Soft Label Generation Methods}
During our experiment for exploring soft label generation, we attempt different CLIP-based models,  evaluating their performances in depicting the similarity between image-text pairs. Our ablative experiment focuses on several representative CLIP-based models \cite{clip,pubmedclip, medclip, pmc-clip,zhang2023biomedclip}. We use the \texttt{base} version of these models to keep all model sizes in a comparative range.

As shown in \cref{tab: ablative soft label generation}, when we focus on the average accuracy, the comparison between \#2 (CLIP \cite{clip}) and the best one \#6 (BioMedCLIP \cite{zhang2023biomedclip}) indicates that soft labels generated from the medical-specific model have better instructional impacts than the model from natural domain. While \#4, \#5, \#6 are all models pre-trained on medical datasets, the average results are close to each other, which demonstrates that the usage of medical knowledge has a nearly equal effect on improving the model representation ability for Med-VQA tasks. In conclusion, we believe a well-trained CLIP-based model in medical domain will benefit the soft label supervision, and thus improving the performance of downstream tasks (\eg, Med-VQA).

\begin{figure}[t]
    \centering
    \includegraphics[width=1\linewidth]{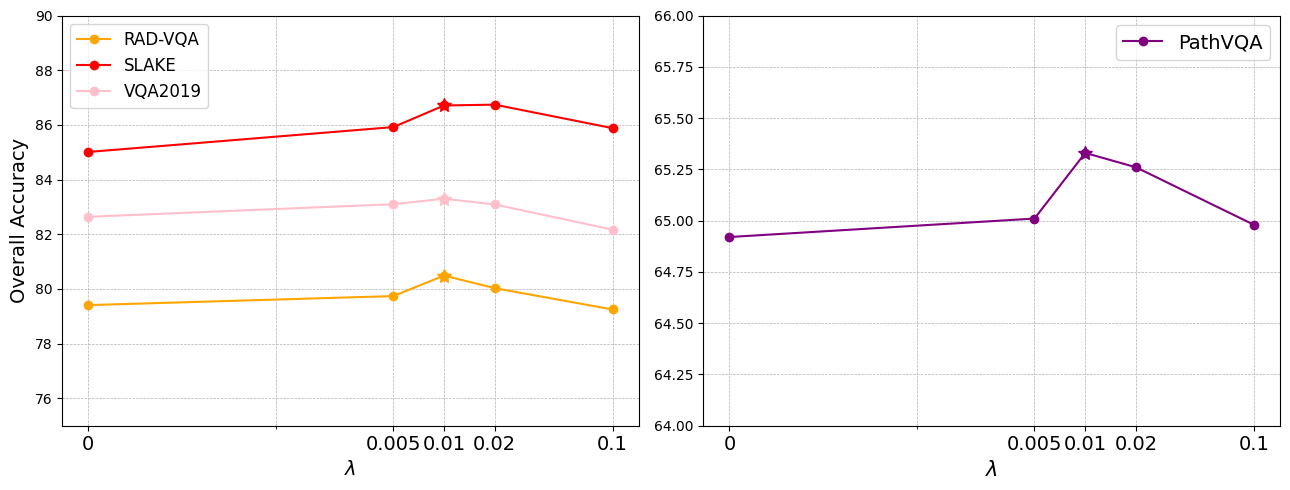}
    \caption{Ablation study on different choices of flexibility weight $\lambda$ for Med-VQA tasks. Datasets are split based on the range of accuracy values.}
    \label{fig: lambda ablative}
\end{figure}

\subsection{Impacts of flexibility weight $\lambda$ for soft labels}
We also evaluate different settings of $\lambda$ that control the positive sample's similarity, and we find out that $\lambda=0.01$ is a locally best choice, as shown in \cref{fig: lambda ablative}. Our explanation for this performance difference is that when we add less weight, the model might ignore the positive samples because of the 
high similarity of both hard negative ones and positive ones. Moreover, if we apply too much weight,  the model will focus too much on positive samples, ignoring the similarity caused by similar diseases among patients, resulting in the decline of performance. Therefore, we believe our setting is a trade-off between focus and ignorance.

%% file: main.bbl
\begin{thebibliography}{59}
\providecommand{\natexlab}[1]{#1}
\providecommand{\url}[1]{\texttt{#1}}
\expandafter\ifx\csname urlstyle\endcsname\relax
  \providecommand{\doi}[1]{doi: #1}\else
  \providecommand{\doi}{doi: \begingroup \urlstyle{rm}\Url}\fi

\bibitem[Andonian et~al.(2022)Andonian, Chen, and Hamid]{andonian2022progressive-self-distillation}
Alex Andonian, Shixing Chen, and Raffay Hamid.
\newblock Robust cross-modal representation learning with progressive self-distillation.
\newblock In \emph{Proceedings of the IEEE/CVF Conference on Computer Vision and Pattern Recognition}, pages 16430--16441, 2022.

\bibitem[Ben~Abacha et~al.(2019)Ben~Abacha, Hasan, Datla, Demner-Fushman, and M{\"u}ller]{data-vqa2019}
Asma Ben~Abacha, Sadid~A Hasan, Vivek~V Datla, Dina Demner-Fushman, and Henning M{\"u}ller.
\newblock Vqa-med: Overview of the medical visual question answering task at imageclef 2019.
\newblock In \emph{Proceedings of CLEF (Conference and Labs of the Evaluation Forum) 2019 Working Notes}. 9-12 September 2019, 2019.

\bibitem[Bodenreider(2004)]{bodenreider2004umls}
Olivier Bodenreider.
\newblock The unified medical language system (umls): integrating biomedical terminology.
\newblock \emph{Nucleic acids research}, 32\penalty0 (suppl\_1):\penalty0 D267--D270, 2004.

\bibitem[Boyd and Vandenberghe(2004)]{boyd2004convex}
Stephen Boyd and Lieven Vandenberghe.
\newblock \emph{Convex optimization}.
\newblock Cambridge university press, 2004.

\bibitem[Chen et~al.(2023{\natexlab{a}})Chen, Zhong, Wu, Luo, and Li]{cmitm}
Cheng Chen, Aoxiao Zhong, Dufan Wu, Jie Luo, and Quanzheng Li.
\newblock Contrastive masked image-text modeling for medical visual representation learning.
\newblock In \emph{International Conference on Medical Image Computing and Computer-Assisted Intervention}, pages 493--503. Springer, 2023{\natexlab{a}}.

\bibitem[Chen et~al.(2020{\natexlab{a}})Chen, Kornblith, Norouzi, and Hinton]{simclr}
Ting Chen, Simon Kornblith, Mohammad Norouzi, and Geoffrey Hinton.
\newblock A simple framework for contrastive learning of visual representations.
\newblock In \emph{International conference on machine learning}, pages 1597--1607. PMLR, 2020{\natexlab{a}}.

\bibitem[Chen et~al.(2023{\natexlab{b}})Chen, He, Xue, Ge, Li, and Yang]{chen2023kobo-rethink-medvlp}
Xiaofei Chen, Yuting He, Cheng Xue, Rongjun Ge, Shuo Li, and Guanyu Yang.
\newblock Knowledge boosting: Rethinking medical contrastive vision-language pre-training.
\newblock In \emph{International Conference on Medical Image Computing and Computer-Assisted Intervention}, pages 405--415. Springer, 2023{\natexlab{b}}.

\bibitem[Chen et~al.(2020{\natexlab{b}})Chen, Li, Yu, El~Kholy, Ahmed, Gan, Cheng, and Liu]{uniter}
Yen-Chun Chen, Linjie Li, Licheng Yu, Ahmed El~Kholy, Faisal Ahmed, Zhe Gan, Yu Cheng, and Jingjing Liu.
\newblock Uniter: Universal image-text representation learning.
\newblock In \emph{European conference on computer vision}, pages 104--120. Springer, 2020{\natexlab{b}}.

\bibitem[Chen et~al.(2022{\natexlab{a}})Chen, Du, Hu, Liu, Li, Wan, and Chang]{m3ae}
Zhihong Chen, Yuhao Du, Jinpeng Hu, Yang Liu, Guanbin Li, Xiang Wan, and Tsung-Hui Chang.
\newblock Multi-modal masked autoencoders for medical vision-and-language pre-training.
\newblock In \emph{International Conference on Medical Image Computing and Computer-Assisted Intervention}, pages 679--689. Springer, 2022{\natexlab{a}}.

\bibitem[Chen et~al.(2022{\natexlab{b}})Chen, Li, and Wan]{arl}
Zhihong Chen, Guanbin Li, and Xiang Wan.
\newblock Align, reason and learn: Enhancing medical vision-and-language pre-training with knowledge.
\newblock In \emph{Proceedings of the 30th ACM International Conference on Multimedia}, pages 5152--5161, 2022{\natexlab{b}}.

\bibitem[Cho et~al.(2021)Cho, Lei, Tan, and Bansal]{cho2021unifying}
Jaemin Cho, Jie Lei, Hao Tan, and Mohit Bansal.
\newblock Unifying vision-and-language tasks via text generation.
\newblock In \emph{International Conference on Machine Learning}, pages 1931--1942. PMLR, 2021.

\bibitem[Dawidowicz et~al.(2023)Dawidowicz, Hirsch, and Tal]{dawidowicz2023limitr}
Gefen Dawidowicz, Elad Hirsch, and Ayellet Tal.
\newblock Limitr: Leveraging local information for medical image-text representation.
\newblock In \emph{Proceedings of the IEEE/CVF International Conference on Computer Vision}, pages 21165--21173, 2023.

\bibitem[Do et~al.(2021)Do, Nguyen, Tjiputra, Tran, Tran, and Nguyen]{mmq2021}
Tuong Do, Binh~X Nguyen, Erman Tjiputra, Minh Tran, Quang~D Tran, and Anh Nguyen.
\newblock Multiple meta-model quantifying for medical visual question answering.
\newblock In \emph{Medical Image Computing and Computer Assisted Intervention--MICCAI 2021: 24th International Conference, Strasbourg, France, September 27--October 1, 2021, Proceedings, Part V 24}, pages 64--74. Springer, 2021.

\bibitem[Dong et~al.(2019)Dong, Yang, Wang, Wei, Liu, Wang, Gao, Zhou, and Hon]{unilm}
Li Dong, Nan Yang, Wenhui Wang, Furu Wei, Xiaodong Liu, Yu Wang, Jianfeng Gao, Ming Zhou, and Hsiao-Wuen Hon.
\newblock Unified language model pre-training for natural language understanding and generation.
\newblock \emph{Advances in neural information processing systems}, 32, 2019.

\bibitem[Eslami et~al.(2023)Eslami, Meinel, and De~Melo]{pubmedclip}
Sedigheh Eslami, Christoph Meinel, and Gerard De~Melo.
\newblock Pubmedclip: How much does clip benefit visual question answering in the medical domain?
\newblock In \emph{Findings of the Association for Computational Linguistics: EACL 2023}, pages 1181--1193, 2023.

\bibitem[Gu et~al.(2024)Gu, Yang, Liu, and Cai]{lapa}
Tiancheng Gu, Kaicheng Yang, Dongnan Liu, and Weidong Cai.
\newblock Lapa: Latent prompt assist model for medical visual question answering.
\newblock In \emph{Proceedings of the IEEE/CVF Conference on Computer Vision and Pattern Recognition (CVPR) Workshops}, 2024.

\bibitem[He et~al.(2024)He, Li, Liu, Zhao, and Zhong]{pefomed}
Jinlong He, Pengfei Li, Gang Liu, Zixu Zhao, and Shenjun Zhong.
\newblock Pefomed: Parameter efficient fine-tuning on multimodal large language models for medical visual question answering.
\newblock \emph{arXiv preprint arXiv:2401.02797}, 2024.

\bibitem[He(2021)]{data-pathvqa}
Xuehai He.
\newblock Towards visual question answering on pathology images.
\newblock In \emph{Proceedings of the 59th annual meeting of the association for computational linguistics and the 11th international joint conference on natural language processing}, 2021.

\bibitem[Hinton et~al.(2015)Hinton, Vinyals, and Dean]{kd}
Geoffrey Hinton, Oriol Vinyals, and Jeff Dean.
\newblock Distilling the knowledge in a neural network.
\newblock \emph{arXiv preprint arXiv:1503.02531}, 2015.

\bibitem[Huang et~al.(2024)Huang, Nie, Wang, and Shang]{huang2024cusa}
Hailang Huang, Zhijie Nie, Ziqiao Wang, and Ziyu Shang.
\newblock Cross-modal and uni-modal soft-label alignment for image-text retrieval.
\newblock In \emph{Proceedings of the AAAI Conference on Artificial Intelligence}, pages 18298--18306, 2024.

\bibitem[Huang et~al.(2019)Huang, Altosaar, and Ranganath]{huang2019clinicalbert}
Kexin Huang, Jaan Altosaar, and Rajesh Ranganath.
\newblock Clinicalbert: Modeling clinical notes and predicting hospital readmission.
\newblock \emph{arXiv preprint arXiv:1904.05342}, 2019.

\bibitem[Huang et~al.(2021)Huang, Shen, Lungren, and Yeung]{gloria}
Shih-Cheng Huang, Liyue Shen, Matthew~P Lungren, and Serena Yeung.
\newblock Gloria: A multimodal global-local representation learning framework for label-efficient medical image recognition.
\newblock In \emph{Proceedings of the IEEE/CVF International Conference on Computer Vision}, pages 3942--3951, 2021.

\bibitem[Jang et~al.(2024)Jang, Kyung, Kim, Lee, Bae, and Choi]{jang2024significantly}
Jongseong Jang, Daeun Kyung, Seung~Hwan Kim, Honglak Lee, Kyunghoon Bae, and Edward Choi.
\newblock Significantly improving zero-shot x-ray pathology classification via fine-tuning pre-trained image-text encoders.
\newblock \emph{Scientific Reports}, 14\penalty0 (1):\penalty0 23199, 2024.

\bibitem[Johnson et~al.(2019)Johnson, Pollard, Greenbaum, Lungren, Deng, Peng, Lu, Mark, Berkowitz, and Horng]{data-mimic}
Alistair~EW Johnson, Tom~J Pollard, Nathaniel~R Greenbaum, Matthew~P Lungren, Chih-ying Deng, Yifan Peng, Zhiyong Lu, Roger~G Mark, Seth~J Berkowitz, and Steven Horng.
\newblock Mimic-cxr-jpg, a large publicly available database of labeled chest radiographs.
\newblock \emph{arXiv preprint arXiv:1901.07042}, 2019.

\bibitem[Khare et~al.(2021)Khare, Bagal, Mathew, Devi, Priyakumar, and Jawahar]{khare2021mmbert}
Yash Khare, Viraj Bagal, Minesh Mathew, Adithi Devi, U~Deva Priyakumar, and CV Jawahar.
\newblock Mmbert: Multimodal bert pretraining for improved medical vqa.
\newblock In \emph{2021 IEEE 18th International Symposium on Biomedical Imaging (ISBI)}, pages 1033--1036. IEEE, 2021.

\bibitem[Kim et~al.(2021)Kim, Son, and Kim]{vilt}
Wonjae Kim, Bokyung Son, and Ildoo Kim.
\newblock Vilt: Vision-and-language transformer without convolution or region supervision.
\newblock In \emph{Proceedings of the 38th International Conference on Machine Learning}, pages 5583--5594. PMLR, 2021.

\bibitem[Lau et~al.(2018)Lau, Gayen, Ben~Abacha, and Demner-Fushman]{data-vqarad}
Jason~J Lau, Soumya Gayen, Asma Ben~Abacha, and Dina Demner-Fushman.
\newblock A dataset of clinically generated visual questions and answers about radiology images.
\newblock \emph{Scientific data}, 5\penalty0 (1):\penalty0 1--10, 2018.

\bibitem[Li et~al.(2021)Li, Selvaraju, Gotmare, Joty, Xiong, and Hoi]{albef}
Junnan Li, Ramprasaath Selvaraju, Akhilesh Gotmare, Shafiq Joty, Caiming Xiong, and Steven Chu~Hong Hoi.
\newblock Align before fuse: Vision and language representation learning with momentum distillation.
\newblock \emph{Advances in neural information processing systems}, 34:\penalty0 9694--9705, 2021.

\bibitem[Li et~al.(2023{\natexlab{a}})Li, Liu, He, Zhao, and Zhong]{mumc}
Pengfei Li, Gang Liu, Jinlong He, Zixu Zhao, and Shenjun Zhong.
\newblock Masked vision and language pre-training with unimodal and multimodal contrastive losses for medical visual question answering.
\newblock In \emph{International Conference on Medical Image Computing and Computer-Assisted Intervention}, pages 374--383. Springer, 2023{\natexlab{a}}.

\bibitem[Li et~al.(2023{\natexlab{b}})Li, Liu, Tan, Liao, and Zhong]{m2i2}
Pengfei Li, Gang Liu, Lin Tan, Jinying Liao, and Shenjun Zhong.
\newblock Self-supervised vision-language pretraining for medial visual question answering.
\newblock In \emph{2023 IEEE 20th International Symposium on Biomedical Imaging (ISBI)}, pages 1--5, 2023{\natexlab{b}}.

\bibitem[Liao et~al.(2021)Liao, Moyer, Cha, Quigley, Berkowitz, Horng, Golland, and Wells]{liao2021local-mutual-information}
Ruizhi Liao, Daniel Moyer, Miriam Cha, Keegan Quigley, Seth Berkowitz, Steven Horng, Polina Golland, and William~M Wells.
\newblock Multimodal representation learning via maximization of local mutual information.
\newblock In \emph{Medical Image Computing and Computer Assisted Intervention--MICCAI 2021: 24th International Conference, Strasbourg, France, September 27--October 1, 2021, Proceedings, Part II 24}, pages 273--283. Springer, 2021.

\bibitem[Lin et~al.(2023{\natexlab{a}})Lin, Zhao, Zhang, Wu, Zhang, Wang, and Xie]{pmc-clip}
Weixiong Lin, Ziheng Zhao, Xiaoman Zhang, Chaoyi Wu, Ya Zhang, Yanfeng Wang, and Weidi Xie.
\newblock Pmc-clip: Contrastive language-image pre-training using biomedical documents.
\newblock In \emph{International Conference on Medical Image Computing and Computer-Assisted Intervention}, pages 525--536. Springer, 2023{\natexlab{a}}.

\bibitem[Lin et~al.(2023{\natexlab{b}})Lin, Bas, Singh, Swaminathan, and Bhotika]{reco}
Zudi Lin, Erhan Bas, Kunwar~Yashraj Singh, Gurumurthy Swaminathan, and Rahul Bhotika.
\newblock Relaxing contrastiveness in multimodal representation learning.
\newblock In \emph{Proceedings of the IEEE/CVF winter conference on applications of computer vision}, pages 2227--2236, 2023{\natexlab{b}}.

\bibitem[Lin et~al.(2023{\natexlab{c}})Lin, Zhang, Tao, Shi, Haffari, Wu, He, and Ge]{lin2023medvqa-survey}
Zhihong Lin, Donghao Zhang, Qingyi Tao, Danli Shi, Gholamreza Haffari, Qi Wu, Mingguang He, and Zongyuan Ge.
\newblock Medical visual question answering: A survey.
\newblock \emph{Artificial Intelligence in Medicine}, 143:\penalty0 102611, 2023{\natexlab{c}}.

\bibitem[Liu et~al.(2021{\natexlab{a}})Liu, Zhan, and Wu]{liu2021cprd}
Bo Liu, Li-Ming Zhan, and Xiao-Ming Wu.
\newblock Contrastive pre-training and representation distillation for medical visual question answering based on radiology images.
\newblock In \emph{Medical Image Computing and Computer Assisted Intervention--MICCAI 2021: 24th International Conference, Strasbourg, France, September 27--October 1, 2021, Proceedings, Part II 24}, pages 210--220. Springer, 2021{\natexlab{a}}.

\bibitem[Liu et~al.(2021{\natexlab{b}})Liu, Zhan, Xu, Ma, Yang, and Wu]{data-slake}
Bo Liu, Li-Ming Zhan, Li Xu, Lin Ma, Yan Yang, and Xiao-Ming Wu.
\newblock Slake: A semantically-labeled knowledge-enhanced dataset for medical visual question answering.
\newblock In \emph{2021 IEEE 18th International Symposium on Biomedical Imaging (ISBI)}, pages 1650--1654. IEEE, 2021{\natexlab{b}}.

\bibitem[Liu et~al.(2022)Liu, Zhan, Xu, and Wu]{cpcr}
Bo Liu, Li-Ming Zhan, Li Xu, and Xiao-Ming Wu.
\newblock Medical visual question answering via conditional reasoning and contrastive learning.
\newblock \emph{IEEE transactions on medical imaging}, 42\penalty0 (5):\penalty0 1532--1545, 2022.

\bibitem[Liu et~al.(2023)Liu, Wang, Xu, and Zhou]{q2atransformer}
Yunyi Liu, Zhanyu Wang, Dong Xu, and Luping Zhou.
\newblock Q2atransformer: Improving medical vqa via an answer querying decoder.
\newblock In \emph{International Conference on Information Processing in Medical Imaging}, pages 445--456. Springer, 2023.

\bibitem[Manmadhan and Kovoor(2020)]{manmadhan2020vqa-survey}
Sruthy Manmadhan and Binsu~C Kovoor.
\newblock Visual question answering: a state-of-the-art review.
\newblock \emph{Artificial Intelligence Review}, 53\penalty0 (8):\penalty0 5705--5745, 2020.

\bibitem[M{\"u}ller et~al.(2022)M{\"u}ller, Kaissis, Zou, and Rueckert]{lovt}
Philip M{\"u}ller, Georgios Kaissis, Congyu Zou, and Daniel Rueckert.
\newblock Joint learning of localized representations from medical images and reports.
\newblock In \emph{European Conference on Computer Vision}, pages 685--701. Springer, 2022.

\bibitem[Nguyen et~al.(2019)Nguyen, Do, Nguyen, Do, Tjiputra, and Tran]{nguyen2019mevf}
Binh~D Nguyen, Thanh-Toan Do, Binh~X Nguyen, Tuong Do, Erman Tjiputra, and Quang~D Tran.
\newblock Overcoming data limitation in medical visual question answering.
\newblock In \emph{Medical Image Computing and Computer Assisted Intervention--MICCAI 2019: 22nd International Conference, Shenzhen, China, October 13--17, 2019, Proceedings, Part IV 22}, pages 522--530. Springer, 2019.

\bibitem[Oord et~al.(2018)Oord, Li, and Vinyals]{loss-infonce}
Aaron van~den Oord, Yazhe Li, and Oriol Vinyals.
\newblock Representation learning with contrastive predictive coding.
\newblock \emph{arXiv preprint arXiv:1807.03748}, 2018.

\bibitem[Pelka et~al.(2018)Pelka, Koitka, R{\"u}ckert, Nensa, and Friedrich]{data-roco}
Obioma Pelka, Sven Koitka, Johannes R{\"u}ckert, Felix Nensa, and Christoph~M Friedrich.
\newblock Radiology objects in context (roco): a multimodal image dataset.
\newblock In \emph{Intravascular Imaging and Computer Assisted Stenting and Large-Scale Annotation of Biomedical Data and Expert Label Synthesis: 7th Joint International Workshop, CVII-STENT 2018 and Third International Workshop, LABELS 2018, Held in Conjunction with MICCAI 2018, Granada, Spain, September 16, 2018, Proceedings 3}, pages 180--189. Springer, 2018.

\bibitem[Peters et~al.(2019)Peters, Neumann, Logan~IV, Schwartz, Joshi, Singh, and Smith]{peters2019knowledge}
Matthew~E Peters, Mark Neumann, Robert~L Logan~IV, Roy Schwartz, Vidur Joshi, Sameer Singh, and Noah~A Smith.
\newblock Knowledge enhanced contextual word representations.
\newblock \emph{arXiv preprint arXiv:1909.04164}, 2019.

\bibitem[Radenovic et~al.(2023)Radenovic, Dubey, Kadian, Mihaylov, Vandenhende, Patel, Wen, Ramanathan, and Mahajan]{radenovic2023Filtering-distillation-hard-negatives}
Filip Radenovic, Abhimanyu Dubey, Abhishek Kadian, Todor Mihaylov, Simon Vandenhende, Yash Patel, Yi Wen, Vignesh Ramanathan, and Dhruv Mahajan.
\newblock Filtering, distillation, and hard negatives for vision-language pre-training.
\newblock In \emph{Proceedings of the IEEE/CVF conference on computer vision and pattern recognition}, pages 6967--6977, 2023.

\bibitem[Radford et~al.(2021)Radford, Kim, Hallacy, Ramesh, Goh, Agarwal, Sastry, Askell, Mishkin, Clark, et~al.]{clip}
Alec Radford, Jong~Wook Kim, Chris Hallacy, Aditya Ramesh, Gabriel Goh, Sandhini Agarwal, Girish Sastry, Amanda Askell, Pamela Mishkin, Jack Clark, et~al.
\newblock Learning transferable visual models from natural language supervision.
\newblock In \emph{International conference on machine learning}, pages 8748--8763. PMLR, 2021.

\bibitem[Rizvi et~al.(2023)Rizvi, Tang, Jiang, Ma, and Hu]{rizvi2023lrclr}
Syed~A Rizvi, Ruixiang Tang, Xiaoqian Jiang, Xiaotian Ma, and Xia Hu.
\newblock Local contrastive learning for medical image recognition.
\newblock In \emph{AMIA Annual Symposium Proceedings}, page 1236. American Medical Informatics Association, 2023.

\bibitem[Shrestha et~al.(2023)Shrestha, Amgain, Khanal, Linte, and Bhattarai]{Shrestha2023-medvlp-review}
Prashant Shrestha, Sanskar Amgain, Bidur Khanal, Cristian~A Linte, and Binod Bhattarai.
\newblock Medical vision language pretraining: A survey.
\newblock \emph{arXiv preprint arXiv:2312.06224}, 2023.

\bibitem[Subramanian et~al.(2020)Subramanian, Wang, Mehta, Bogin, van Zuylen, Parasa, Singh, Gardner, and Hajishirzi]{data-medicat}
Sanjay Subramanian, Lucy~Lu Wang, Sachin Mehta, Ben Bogin, Madeleine van Zuylen, Sravanthi Parasa, Sameer Singh, Matt Gardner, and Hannaneh Hajishirzi.
\newblock Medicat: A dataset of medical images, captions, and textual references.
\newblock \emph{arXiv preprint arXiv:2010.06000}, 2020.

\bibitem[Tanida et~al.(2023)Tanida, M{\"u}ller, Kaissis, and Rueckert]{tanida2023interactive}
Tim Tanida, Philip M{\"u}ller, Georgios Kaissis, and Daniel Rueckert.
\newblock Interactive and explainable region-guided radiology report generation.
\newblock In \emph{Proceedings of the IEEE/CVF Conference on Computer Vision and Pattern Recognition}, pages 7433--7442, 2023.

\bibitem[Veli{\v{c}}kovi{\'c} et~al.(2017)Veli{\v{c}}kovi{\'c}, Cucurull, Casanova, Romero, Lio, and Bengio]{velivckovic2017graph-attention-network}
Petar Veli{\v{c}}kovi{\'c}, Guillem Cucurull, Arantxa Casanova, Adriana Romero, Pietro Lio, and Yoshua Bengio.
\newblock Graph attention networks.
\newblock \emph{arXiv preprint arXiv:1710.10903}, 2017.

\bibitem[Wang et~al.(2022{\natexlab{a}})Wang, Zhou, Wang, Vardhanabhuti, and Yu]{mgca}
Fuying Wang, Yuyin Zhou, Shujun Wang, Varut Vardhanabhuti, and Lequan Yu.
\newblock Multi-granularity cross-modal alignment for generalized medical visual representation learning.
\newblock \emph{Advances in Neural Information Processing Systems}, 35:\penalty0 33536--33549, 2022{\natexlab{a}}.

\bibitem[Wang et~al.(2021)Wang, Gao, Zhu, Zhang, Liu, Li, and Tang]{wang2021kepler}
Xiaozhi Wang, Tianyu Gao, Zhaocheng Zhu, Zhengyan Zhang, Zhiyuan Liu, Juanzi Li, and Jian Tang.
\newblock Kepler: A unified model for knowledge embedding and pre-trained language representation.
\newblock \emph{Transactions of the Association for Computational Linguistics}, 9:\penalty0 176--194, 2021.

\bibitem[Wang et~al.(2022{\natexlab{b}})Wang, Wu, Agarwal, and Sun]{medclip}
Zifeng Wang, Zhenbang Wu, Dinesh Agarwal, and Jimeng Sun.
\newblock Medclip: Contrastive learning from unpaired medical images and text.
\newblock \emph{arXiv preprint arXiv:2210.10163}, 2022{\natexlab{b}}.

\bibitem[Xie et~al.(2020)Xie, Wang, Wang, and Zha]{ipot-algorithm}
Yujia Xie, Xiangfeng Wang, Ruijia Wang, and Hongyuan Zha.
\newblock A fast proximal point method for computing exact wasserstein distance.
\newblock In \emph{Uncertainty in artificial intelligence}, pages 433--453. PMLR, 2020.

\bibitem[Yuan et~al.(2023)Yuan, Jin, Tan, Zhao, Yuan, Huang, and Huang]{yuan2023ramm}
Zheng Yuan, Qiao Jin, Chuanqi Tan, Zhengyun Zhao, Hongyi Yuan, Fei Huang, and Songfang Huang.
\newblock Ramm: Retrieval-augmented biomedical visual question answering with multi-modal pre-training.
\newblock In \emph{Proceedings of the 31st ACM International Conference on Multimedia}, pages 547--556, 2023.

\bibitem[Zhang et~al.(2023)Zhang, Xu, Usuyama, Xu, Bagga, Tinn, Preston, Rao, Wei, Valluri, et~al.]{zhang2023biomedclip}
Sheng Zhang, Yanbo Xu, Naoto Usuyama, Hanwen Xu, Jaspreet Bagga, Robert Tinn, Sam Preston, Rajesh Rao, Mu Wei, Naveen Valluri, et~al.
\newblock Biomedclip: a multimodal biomedical foundation model pretrained from fifteen million scientific image-text pairs.
\newblock \emph{arXiv preprint arXiv:2303.00915}, 2023.

\bibitem[Zhang et~al.(2022)Zhang, Jiang, Miura, Manning, and Langlotz]{convirt}
Yuhao Zhang, Hang Jiang, Yasuhide Miura, Christopher~D Manning, and Curtis~P Langlotz.
\newblock Contrastive learning of medical visual representations from paired images and text.
\newblock In \emph{Machine Learning for Healthcare Conference}, pages 2--25. PMLR, 2022.

\bibitem[Zhou et~al.(2023)Zhou, Lian, Wang, and Yu]{mrm-iclr}
Hong-Yu Zhou, Chenyu Lian, Liansheng Wang, and Yizhou Yu.
\newblock Advancing radiograph representation learning with masked record modeling.
\newblock \emph{arXiv preprint arXiv:2301.13155}, 2023.

\end{thebibliography}
